\documentclass[conference, letterpaper]{IEEEtran}
\IEEEoverridecommandlockouts

\usepackage{cite}
\usepackage{amsmath,amssymb,amsfonts}
\usepackage{algorithmic}
\usepackage{graphicx}
\usepackage{textcomp}
\usepackage{multirow}
\usepackage{multicol}
\usepackage{subfigure}
\usepackage{xcolor}
\usepackage[letterpaper, right=1.7cm, left=1.7cm, top=1.9cm, bottom=3.45cm]{geometry}
\def\BibTeX{{\rm B\kern-.05em{\sc i\kern-.025em b}\kern-.08em
    T\kern-.1667em\lower.7ex\hbox{E}\kern-.125emX}}

\begin{document}
\title{Joint Memory Frequency and Computing Frequency Scaling for Energy-efficient DNN Inference}
\author{\IEEEauthorblockN{
Yunchu~Han,~Zhaojun~Nan,~Sheng~Zhou,~and~Zhisheng~Niu
}
\IEEEauthorblockA{
Beijing National Research Center for Information Science and Technology\\Department of Electronic Engineering, Tsinghua University, Beijing 100084, China}
Emails: \{hyc23@mails.,~nzj660624@mail.,~sheng.zhou@,~niuzhs@\}tsinghua.edu.cn}
\maketitle

\begin{abstract}
Deep neural networks (DNNs) have been widely applied in diverse applications, but the problems of high latency and energy overhead are inevitable on resource-constrained devices. To address this challenge, most researchers focus on the dynamic voltage and frequency scaling (DVFS) technique to balance the latency and energy consumption by changing the computing frequency of processors. However, the adjustment of memory frequency is usually ignored and not fully utilized to achieve efficient DNN inference, which also plays a significant role in the inference time and energy consumption. In this paper, we first investigate the impact of joint memory frequency and computing frequency scaling on the inference time and energy consumption with a model-based and data-driven method. Then by combining with the fitting parameters of different DNN models, we give a preliminary analysis for the proposed model to see the effects of adjusting memory frequency and computing frequency simultaneously. Finally, simulation results in local inference and cooperative inference cases further validate the effectiveness of jointly scaling the memory frequency and computing frequency to reduce the energy consumption of devices.
\end{abstract}

\section{Introduction}\label{sec:intro}
Towards the ubiquitous applications in 6G, artificial intelligence (AI) technology has increasingly played a crucial role \cite{WangC}. Although deep neural networks (DNNs) have been leveraged in various applications (e.g., image classification, object detection, intelligent content generation), their complicated architectures and multiple parameters lead to substantial computation overhead, intolerable latency, and high energy consumption \cite{Sze}. Edge intelligence \cite{ZhouZ} has been recognized as a promising solution to these problems by exploiting mobile edge computing (MEC) \cite{MaoY} to provide intelligent services at the edge of mobile networks. In addition, dynamic voltage and frequency scaling (DVFS) \cite{Rabaey} technique can also be applied to optimize the computing frequency of processors for timely and energy-efficient DNN inference. Various studies have focused on the combination of DVFS with different aspects, such as DNN partitioning \cite{NanZRobustPartition}, batch processing \cite{Nabavinejad}, \cite{ShiW}, task offloading and resource allocation \cite{NanRobustOffload}, \cite{NanOffloadFeedback}. 

However, in addition to optimizing the computing frequency (e.g., CPU frequency, GPU frequency), scaling the memory frequency (e.g., DRAM frequency, embedded memory controller frequency) is also essential to improve the latency and energy consumption of DNN inference. Specifically, \cite{ZhangZ} optimizes the computing frequency and memory frequency of edge devices as well as the offloading proportion through deep reinforcement learning to reduce the energy consumption and end-to-end latency. The techniques of quantization and memory voltage scaling are explored in \cite{Denkinger} to reduce the memory footprint, where the corresponding performance and energy consumption are analyzed. In \cite{Parmar}, memory-oriented hardware design and optimization are introduced to achieve energy conservation in extended reality applications, and the energy proportion for memory access and computation is analyzed. The authors in \cite{WangQ} focus on the impact of computing frequency and memory frequency scaling on the execution time of GPU kernels, and a fine-grained analytical model is revealed. Therefore, it is a potential approach to achieve efficient DNN inference with joint optimization of memory frequency and other factors. 
\begin{figure}[t]
  \centering
  \includegraphics[width=0.4\textwidth]{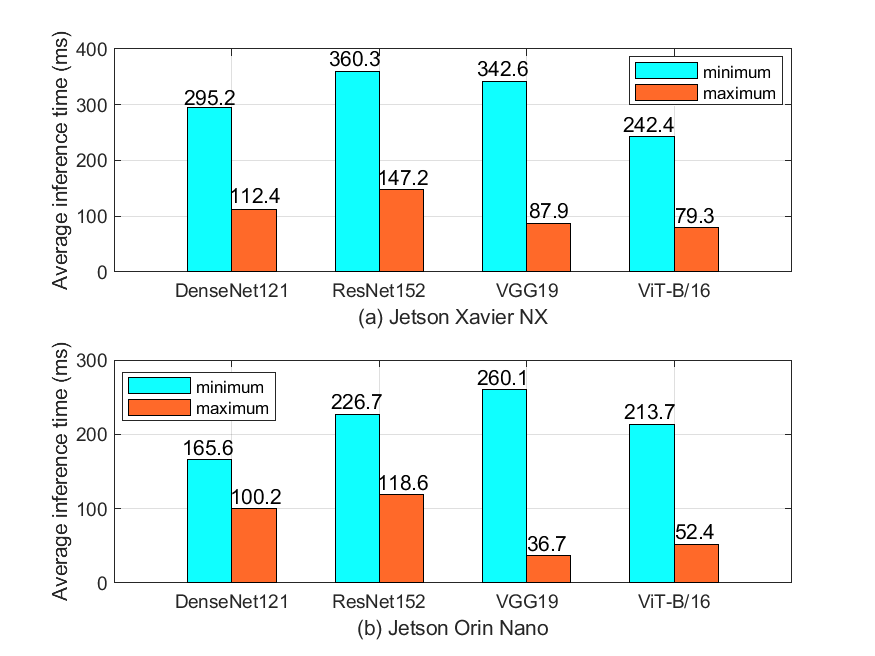}\\
  \caption{Average inference time of different DNNs on Jetson Xavier NX and Jetson Orin Nano when GPU frequency is fixed to the maximum.}
  \label{fig:myFig1}
  \vspace{-0.3cm}
\end{figure}

However, these existing studies do not provide an appropriate formulation for the impact of both memory frequency and computing frequency on the latency and energy consumption of DNN inference. The analytical model proposed in \cite{WangQ} does not directly give the impact of memory frequency, and there is an important difference between kernel-based applications and DNN inference tasks. To clearly present the impact of memory frequency scaling on DNN inference time, we keep the computing frequency constant and evaluate the average inference time of four typical DNNs (i.e., DenseNet121 \cite{HuangG}, ResNet152 \cite{HeK}, VGG19 \cite{Simonyan}, and ViT-B/16 \cite{Dosovitskiy}) with minimum and maximum memory frequency, respectively. From Fig. \ref{fig:myFig1}, we observe that there is a significant decline in the average inference time with memory frequency scaling, which varies for different DNN models and devices. For example, we can achieve an average reduction of $74\%$ for VGG19 but only $59\%$ for ResNet152 on Jetson Xavier NX, while the corresponding reductions are $86\%$ and $48\%$ on Jetson Orin Nano. Therefore, it is necessary to analyze the impact of joint memory frequency and computing frequency scaling on the latency and energy consumption of DNN inference.

In this work, joint scaling of memory frequency and computing frequency is introduced to achieve energy-efficient DNN inference. We first characterize the impact of scaling both memory frequency and computing frequency on DNN inference time based on the DVFS technique. Then by combining with the real-world data and model-based method, the average inference time is formulated as a function related to the memory frequency and computing frequency. Furthermore, a three-dimensional figure of the latency and energy consumption is presented for intuitive understanding and analysis. Finally, we evaluate the effects of joint memory frequency and computing frequency scaling in local inference and edge-device cooperative inference cases, where the energy consumption can be reduced by jointly scaling these two frequencies.

\section{System Overview}\label{sec:system}
\subsection{System Model}
\begin{figure}[t]
  \centering
  \includegraphics[width=0.4\textwidth]{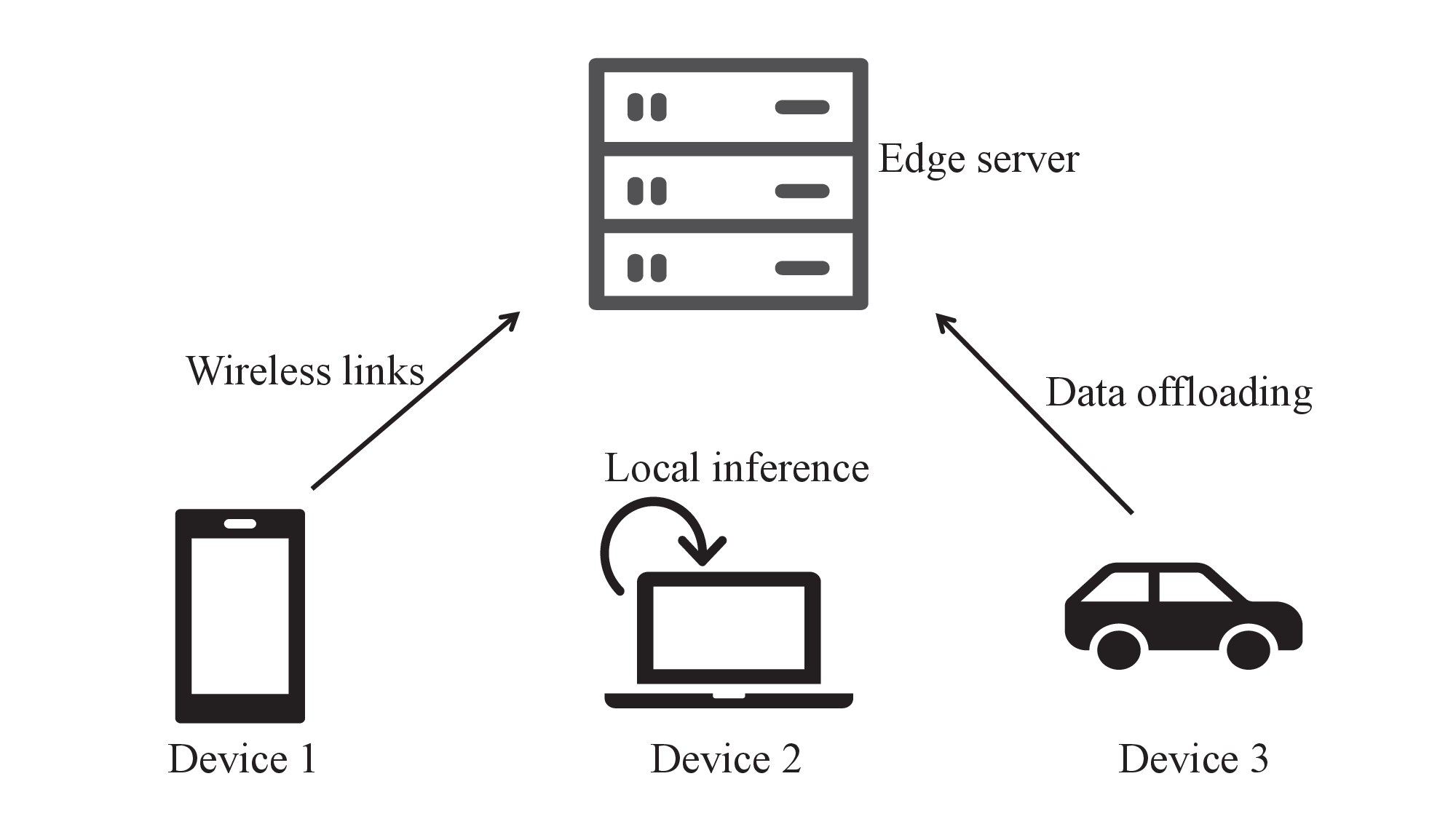}\\
  \caption{\leftline{Illustration of an edge intelligence system.}}
  \label{fig:myFig2}
  \vspace{-0.3cm}
\end{figure}

As illustrated in Fig. \ref{fig:myFig2}, an edge intelligence system consisting of $N$ mobile devices (denoted as $\mathcal{N} \triangleq \{1, 2, \dots, N\}$) and an edge server is considered. We assume that each device has deployed pre-trained DNNs (e.g., VGG, ResNet, and ViT) to complete inference tasks. The binary offloading policy $x_n \in \{0, 1\}$ is adopted, where $x_n = 0$ means that device $n$ executes local inference, and $x_n = 1$ means that device $n$ offloads the task to the edge server and executes edge inference. The result feedback latency is ignored in this work because of its relatively small data volume. 

\subsection{Inference Time}
Denote $c_n$ and $f_n$ as the computation workload (in floating-point operations, i.e. FLOPs) and the computing frequency of device $n$ (in cycle/s), respectively. To formulate the impact of computing frequency on inference time, a widely-used model is 
\begin{align}\label{eq:myeq1}
    t_{n} = \frac{c_n}{g_n f_n}, \forall n \in \mathcal{N},
\end{align}
where $g_n$ means the number of FLOPs per cycle, and it mainly depends on the type of processor. However, this formulation exists several defects. Firstly, there is an inconsistent relationship between inference time and required FLOPs under the fixed computing frequency of devices \cite{Han}. Secondly, as shown in Fig. \ref{fig:myFig1}, changing the memory frequency of devices makes a significant difference in the inference time, but there are few studies that focus on the effect of memory frequency. Finally, since both computing frequency and memory frequency can influence the inference time, the optimization results may be variable in different situations. For example, a trade-off can manifest in scenarios where devices with constrained computing capacity but high memory frequency can achieve comparable inference performance to devices featuring superior computing capacity but limited memory frequency. Therefore, it is necessary to jointly consider the impact of memory frequency and computing frequency on inference time and energy consumption, and the corresponding details are given in Section \ref{sec:model}.  

\subsection{Energy Consumption}
\begin{figure}[t]
  \centering
  \includegraphics[width=0.4\textwidth]{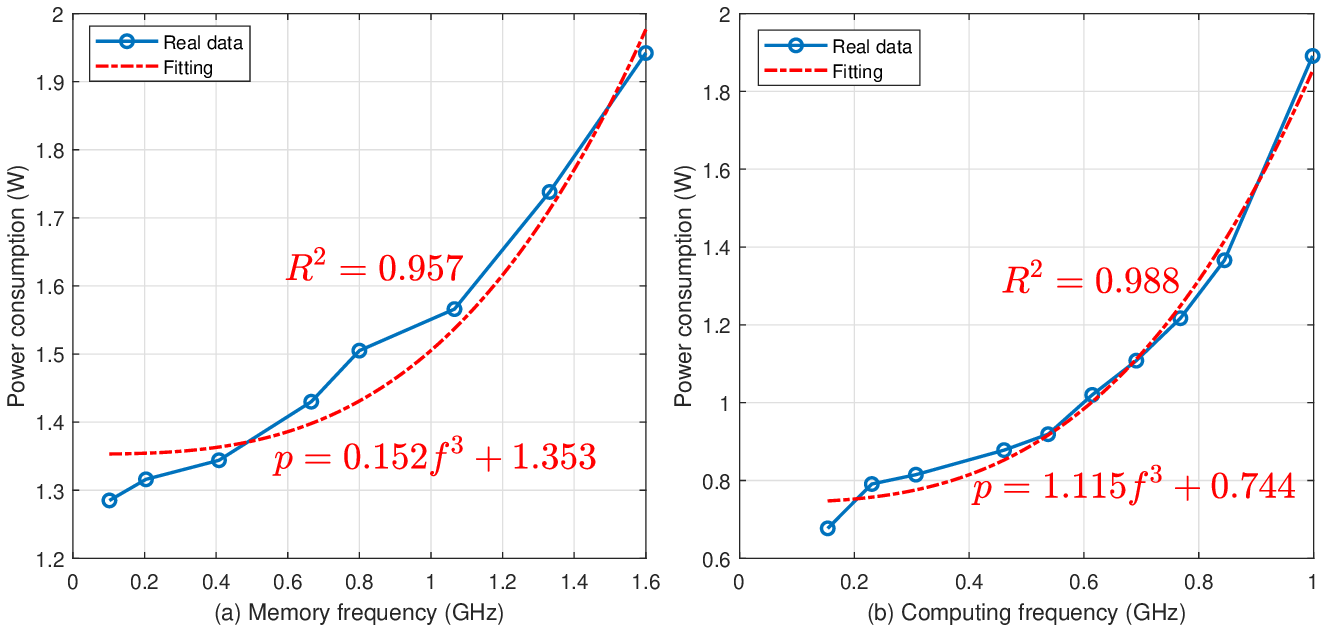}\\
  \caption{Average power consumption characteristics of memory frequency and computing frequency scaling on Jetson TX1.}
  \label{fig:myFig3}
  \vspace{-0.3cm}
\end{figure}

\begin{figure}[t]
  \centering
  \includegraphics[width=0.4\textwidth]{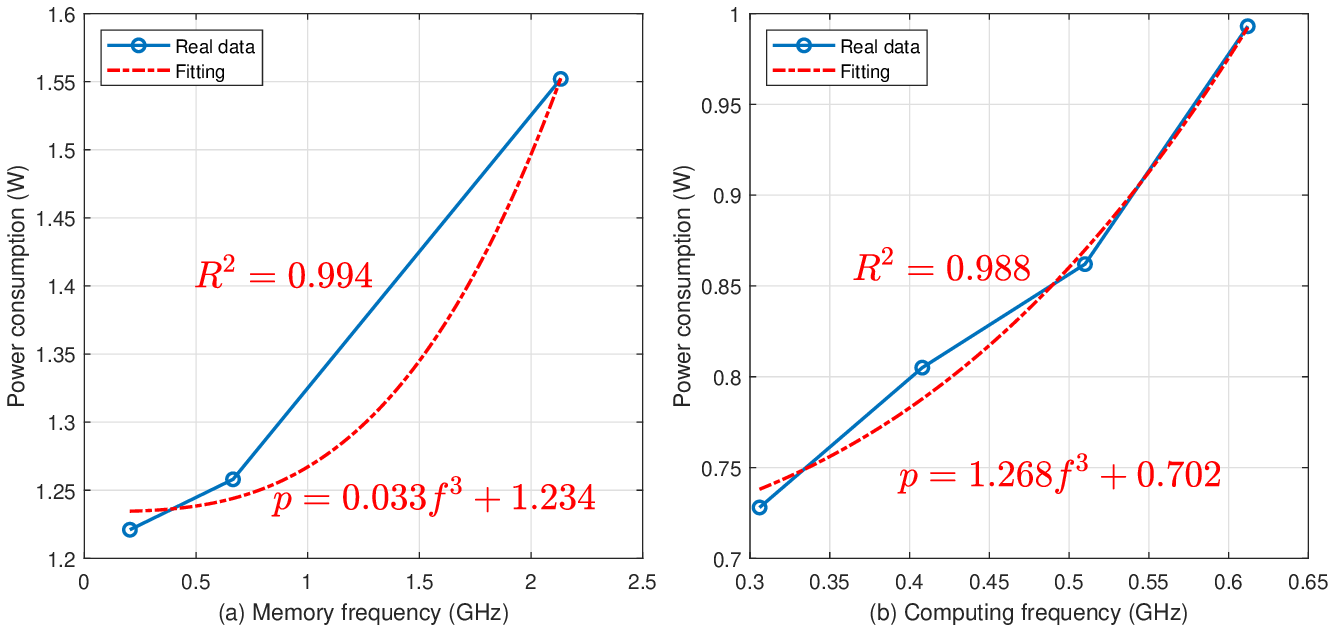}\\
  \caption{Average power consumption characteristics of memory frequency and computing frequency scaling on Jetson Orin Nano.}
  \label{fig:myFig4}
  \vspace{-0.3cm}
\end{figure}

To analyze the energy consumption, it is essential to obtain the power characteristics of memory frequency and computing frequency scaling. A general formulation for the dynamic power of CMOS circuit is given by $P = \alpha C V^2 f$, where $\alpha$, $C$, $V$ and $f$ denote the activity factor, the capacity, the supply voltage and the frequency, respectively \cite{Haj}. However, the power consumption caused by frequent memory access is usually ignored. To further investigate the impact of memory frequency and computing frequency scaling on power consumption, we conduct experiments on NVIDIA Jetson TX1 and Jetson Orin Nano, where we independently vary one frequency while maintaining the other frequency constant. The real data and fitting results are shown in Fig. \ref{fig:myFig3} and Fig. \ref{fig:myFig4}, where the correlation coefficients are all larger than $0.95$. When the computing frequency is fixed, the power consumption of memory access approximately shows a cubic relationship with the memory frequency, and the power consumption of computation also shows a similar trend with the computing frequency. According to the fitting parameters, we observe that the impact of computing frequency scaling on power consumption is stronger than the impact of memory frequency scaling. It should be emphasized that there is relatively sparse data for Jetson Orin Nano since the available computing and memory frequencies are finite. Based on the fitting results, the power consumption of device $n$ for local inference is given by
\begin{align}\label{eq:myeq2}
    p_n = \kappa_{n, \rm{mem}} f_{n, \rm{mem}}^3 + \kappa_{n, \rm{com}} f_{n, \rm{com}}^3 + \sigma_n, \forall n \in \mathcal{N},
\end{align}
where $\kappa_{n, \rm{mem}}$ and $\kappa_{n, \rm{com}}$ denote the equivalent power coefficients, $f_{n, \rm{mem}}$ and $f_{n, \rm{com}}$ denote the memory frequency and computing frequency of device $n$, and $\sigma_n$ is a constant term. Therefore, the energy consumption of device $n$ is given by
\begin{align}\label{eq:myeq3}
    e_n = p_n t_n, \forall n \in \mathcal{N}.
\end{align}

\emph{Remark 1:} Compared with other related work formulating the power consumption of computation as $ P = \kappa f^3$, we add a constant term to better satisfy the real data. Furthermore, we apply this model to describe the power consumption caused by the memory frequency scaling, and the impacts of these two frequencies on power consumption are independent. 

\section{Modeling and Analysis}\label{sec:model}
\subsection{Latency Modeling Based on Joint Memory Frequency and Computing Frequency Scaling}
\begin{figure}[t]
    \centering
    \subfigure[ResNet152 on Jetson TX1]{\includegraphics[width=0.35\textwidth]{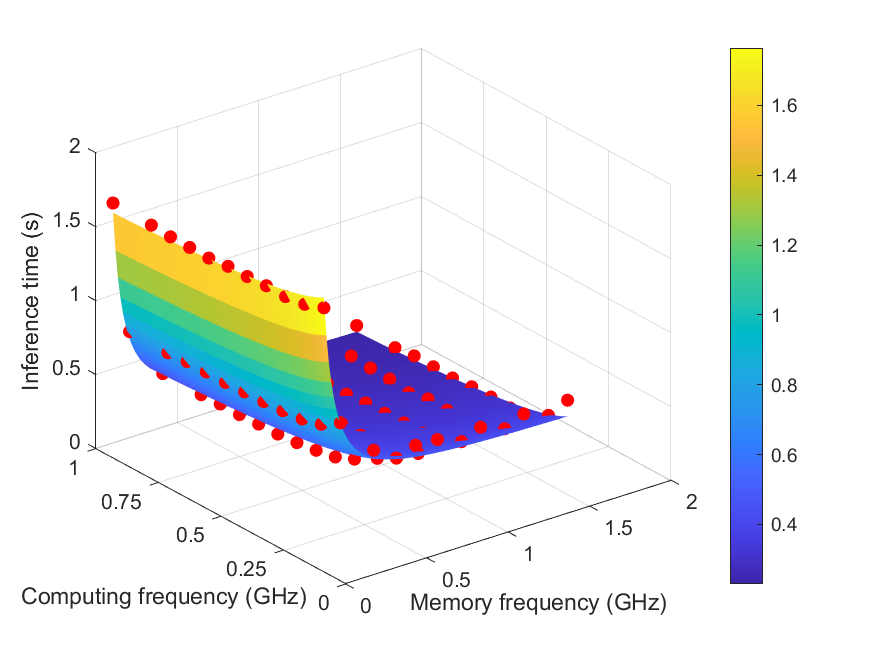}
    \label{fig:myFig5a}}
    \subfigure[ViT-B/16 on Jetson Orin Nano]{\includegraphics[width=0.35\textwidth]{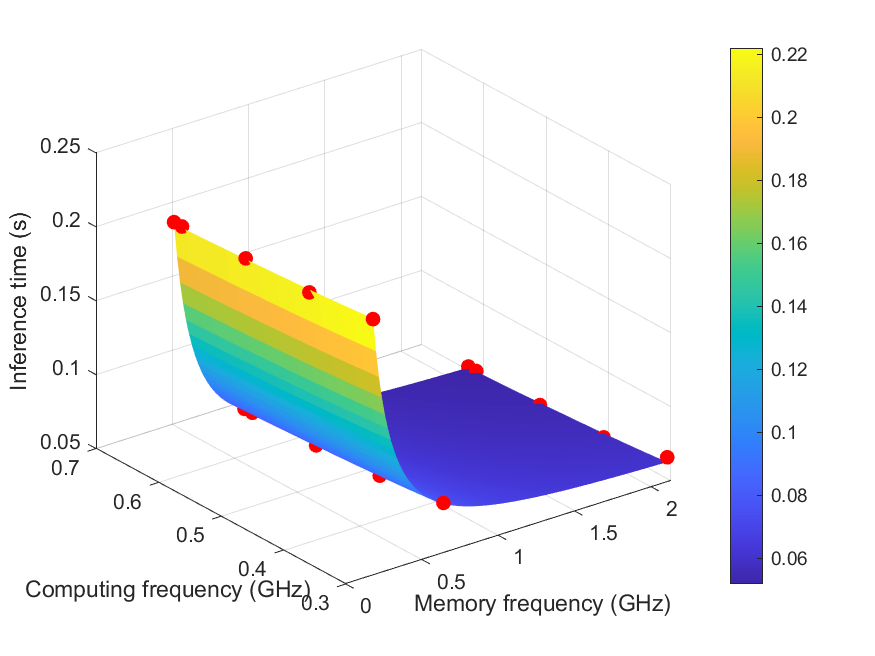}
    \label{fig:myFig5b}}
    \caption{Visulization of the latency models for ResNet152 and ViT-B/16.}
    \label{fig:myFig5}
    \vspace{-0.3cm}
\end{figure}

It is known that devices with higher computing frequency can achieve faster inference. However, there is a marginal effect when the computing frequency reaches a threshold. In addition, with the increasing of DNN model size and parameters, data transmission efficiency gradually plays a dominant role in DNN inference time. To evaluate the impact of joint memory frequency and computing frequency scaling on DNN inference time, we measure the average inference time of four representative DNNs on Jetson TX1 and Orin Nano by configuring different combinations of these two frequencies. Since there are two independent variables affecting DNN inference time, we give two examples for intuitive visualization, as shown in Fig. \ref{fig:myFig5a} and Fig. \ref{fig:myFig5b}. Based on the real-world data and empirical formulation in (\ref{eq:myeq1}), the impact of memory frequency and computing frequency on DNN inference time can be formulated by
\begin{equation}
    t_n = \lambda_n f_{n, \rm{mem}}^{-\beta_n} + \mu_n f_{n, \rm{com}}^{-\gamma_n}, \forall n \in \mathcal{N},
\end{equation}
where $\lambda_n$, $\beta_n$, $\mu_n$ and $\gamma_n$ are all positive fitting parameters relevant to the hardware architecture, the type of DNNs, etc. The memory frequency or computing frequency is in Giga Hz, and the inference time is in second. The parameters $\lambda_n$ and $\beta_n$ mainly reflect the effect of memory frequency, while $\mu_n$ and $\gamma_n$ show the impact of computing frequency. The corresponding parameters for these DNNs and platforms are shown in Table \ref{tab:mytab1}. The mean squared error (MSE) is smaller than $0.002$ and R-squared ($\rm{R^2}$) is larger than 0.99, indicating the precision of these fitting results. Due to the weak computing capacity of Jetson TX1, we do not execute inference with ViT-B/16 on it.
\begin{table*}[!htbp]
    \centering
    \caption{Fitting parameters of different DNNs on Jetson TX1 and Orin Nano.}
    \begin{tabular}{c|ccc|ccccc}
    \hline
    \multirow{2}*{\textbf{Parameters}} & \multicolumn{3}{c|}{\textbf{Jetson TX1}} & \multicolumn{4}{c}{\textbf{Jetson Orin Nano}} \\
    \cline{2-8}
    \multirow{2}*{} & \textbf{VGG19} & \textbf{ResNet152} & \textbf{DenseNet121} & \textbf{VGG19} & \textbf{ResNet152} & \textbf{DenseNet121} & \textbf{ViT-B/16} \\
    \hline
    $\lambda_n$ & 0.110 & 0.039   & 0.010             & 0.023  & 0.012  & 0.013  & 0.012 \\
    \hline
    $\beta_n$   & 1.296 & 1.552   & 1.664             & 1.416  & 1.426  & 1.040  & 1.638  \\
    \hline
    $\mu_n$     & 0.099 & 0.211   & 0.172             & 0.026  & 0.112  & 0.096  & 0.043 \\
    \hline
    $\gamma_n$  & 0.629 & 0.358   & 0.082             & 0.555  & 0.036  & 0.017  & 0.241 \\
    \hline
    $\rm{R^2}$  & 0.9961 & 0.9917 & 0.9967            & 0.9974 & 0.9993 & 0.9950 & 0.9996 \\
    \hline
    MSE         & 0.0017 & 0.0016 & 0.0001            & 2e-5   & 2e-6   & 3e-6   & 2e-6 \\
    \hline
    \end{tabular}
    \label{tab:mytab1}
    \vspace{-0.25cm}
\end{table*}

\subsection{Analysis for the Proposed Model}
Although it is a fact that changing memory frequency can affect DNN inference time, there is no formal mathematical formulation. Our proposed model first gives one way to demonstrate the impact of both memory frequency and computing frequency, and further optimization can be achieved since the function is convex. The required computation and memory access costs vary for different DNN models, and there is no obvious relationship between these costs and the fitting parameters. However, it seems that the impact of memory frequency scaling is more significant than computing frequency because the parameter $\beta_n$ is larger than $\gamma_n$ for these DNN models. In addition, the heterogeneous computing and memory access capacity of devices should be taken into account. For example, DenseNet121 needs much fewer FLOPs and memory access costs than ResNet152, and the fitting parameters of DenseNet121 are close to or smaller than those of ResNet152 for Jetson Orin Nano. However, the similar trend is not observed for Jetson TX1 because the parameter $\lambda_n$ is much smaller and $\beta_n$ is larger than ResNet152. 

\begin{figure}[!htbp]
  \centering
  \subfigure[ResNet152 on Jetson TX1]{\includegraphics[width=0.35\textwidth]{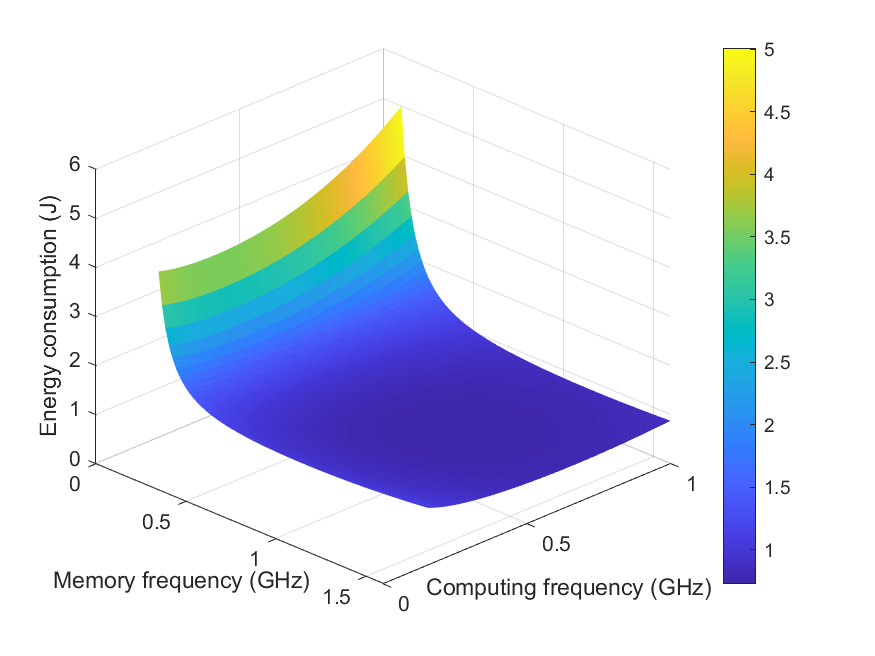}
  \label{fig:myFig6a}}
  \subfigure[ViT-B/16 on Jetson Orin Nano]{\includegraphics[width=0.35\textwidth]{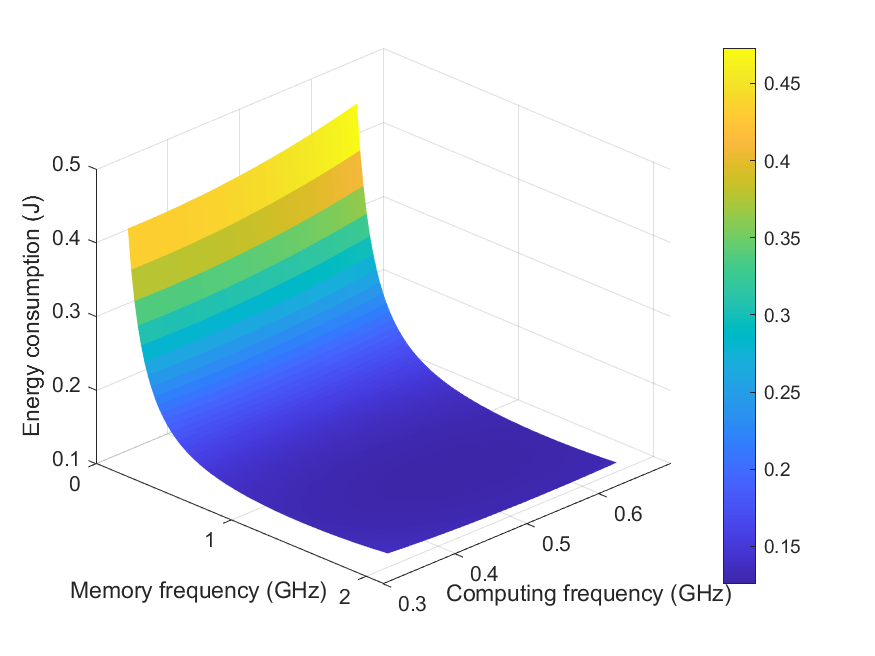}
  \label{fig:myFig6b}}
  \caption{Energy consumption with joint memory frequency and memory frequency scaling for ResNet152 and ViT-B/16.}
  \label{fig:myFig6}
  \vspace{-0.25cm}
\end{figure}

Given the proposed latency model, it is necessary to analyze the corresponding energy consumption to achieve energy-efficient DNN inference. As shown in Fig. \ref{fig:myFig6}, we give an illustration for the energy consumption of ResNet152 on Jetson TX1 and ViT-B/16 on Jetson Orin Nano, respectively. When the computing frequency is fixed, the energy consumption shows a descending trend as the memory frequency increases. In this situation, the decreasing rate of inference time is much larger than the increasing speed of power consumption. 
In Fig. \ref{fig:myFig6a}, the energy consumption also decreases monotonously when we decrease the computing frequency with the memory frequency remaining at a relatively small value. However, this relationship no longer holds if the memory frequency stays at a higher value. In Fig. \ref{fig:myFig6b}, we observe that the impact of memory frequency scaling on the energy consumption is more significant than the impact of computing frequency, which is consistent with the analysis in \ref{sec:system} and \ref{sec:model}. Since the characteristics of energy consumption with joint memory frequency and computing frequency scaling depend on multiple factors (e.g., the type of DNN models, hardware, and fitting parameters), we should give the corresponding analysis based on the specific scenarios.

\section{Simulation Results}\label{sec:eval}
\subsection{Local Inference via Frequency Scaling}
We first consider a universal case where devices execute DNN inference locally. VGG19 and DenseNet121 are deployed on Jetson TX1 to execute image classification tasks on the CIFAR-100 dataset \cite{Kriz}, while ResNet152 and ViT-B/16 are selected on Jetson Orin Nano. The detailed configuration for the memory frequency and computing frequency range is shown in Table \ref{tab:mytab2}. We consider three possible policies for adjusting memory frequency and computing frequency to meet the deadline constraint. 
\begin{enumerate}
    \item \textbf{Computing-prior policy:} Fix the computing frequency to the maximum value and adjust the memory frequency dynamically.
    \item \textbf{Memory-prior policy:} Fix the memory frequency to the maximum value and adjust the computing frequency dynamically.
    \item \textbf{Joint scaling policy:} Simultaneously adjust the memory frequency and the computing frequency.
\end{enumerate}
\begin{table}[!htbp]
    \centering
    \caption{Configuration of devices.}
    \begin{tabular}{c|c|c}
    \hline
    \textbf{Device}          &  \textbf{Memory frequency}    &  \textbf{Computing frequency}    \\
    \hline
    Jetson TX1      &  [0.102, 1.6] GHz    &  [0.1536, 0.9984] GHz   \\
    \hline
    Jetson Orin Nano &  [0.204, 2.133] GHz  &  [0.306, 0.62475] GHz   \\
    \hline
    \end{tabular}
    \label{tab:mytab2}
\end{table}

\begin{figure}[!htbp]
  \centering
  \includegraphics[width=0.4\textwidth]{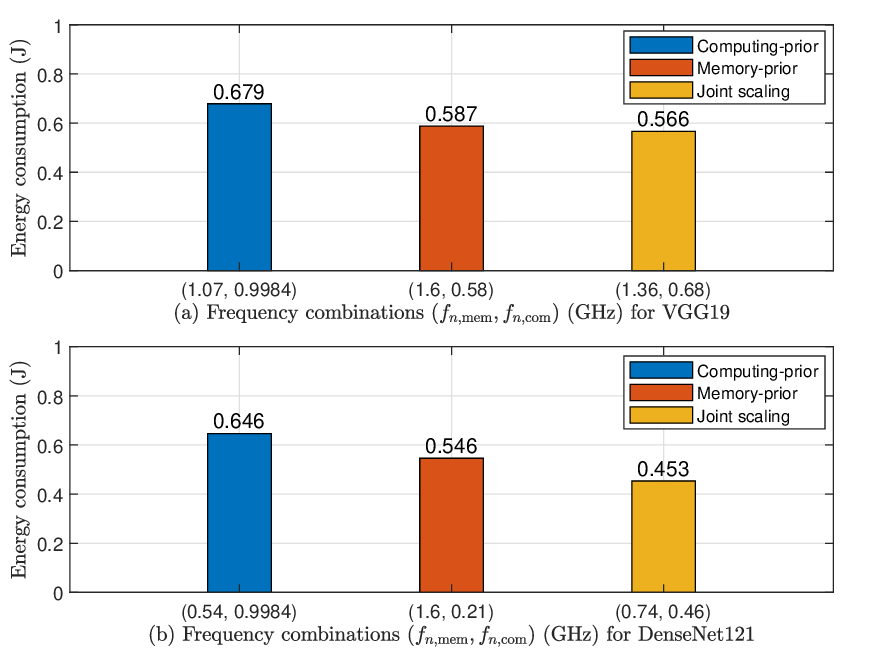}\\
  \caption{Energy consumption of three frequency scaling policies for VGG19 and DenseNet121 on Jetson TX1.}
  \label{fig:myFig7}
\end{figure}

Fig. \ref{fig:myFig7} shows the impact of the three proposed frequency scaling policies on the energy consumption of VGG19 and DenseNet121 with the given deadline $D = 0.2$ s on Jetson TX1. There is a trade-off between memory frequency and computing frequency, which is consistent with the proposed models shown in Fig. \ref{fig:myFig5}. The memory frequency can be significantly reduced when adopting the computing-prior policy, but the energy consumption shows no advantages. Similarly, the energy consumption is still high if we use the memory-prior policy. However, by jointly scaling both memory frequency and computing frequency, we can achieve an average reduction of $10\%$ and $23\%$ in the energy consumption for VGG19 and DenseNet121, respectively. 

\begin{figure}[!htbp]
  \centering
  \includegraphics[width=0.4\textwidth]{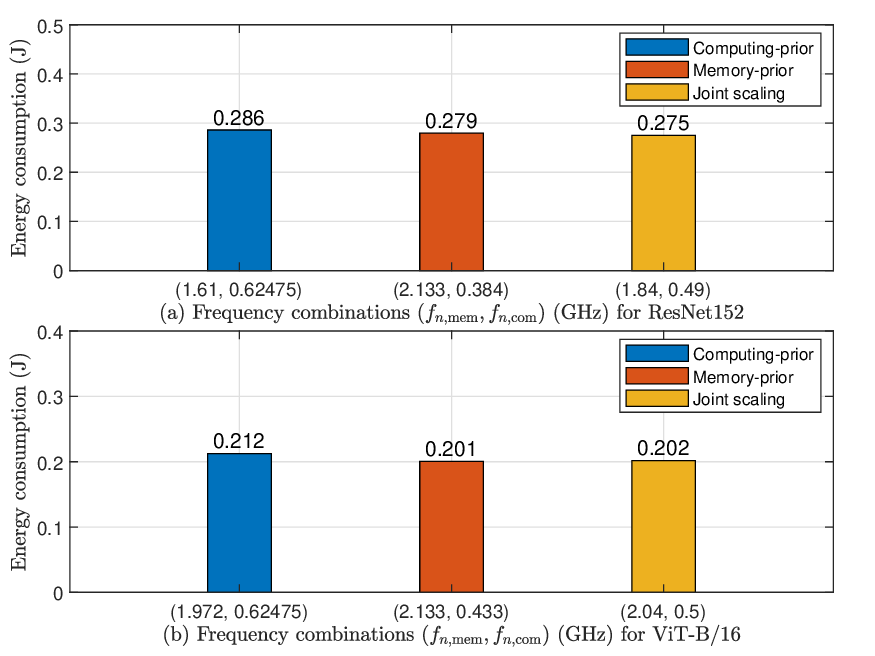}\\
  \caption{Energy consumption of three frequency scaling policies for ResNet152 and ViT-B/16 on Jetson Orin Nano.}
  \label{fig:myFig8}
  \vspace{-0.25cm}
\end{figure}

In Fig. \ref{fig:myFig8}, we evaluate the corresponding energy consumption for ResNet152 and ViT-B/16 on Jetson Orin Nano, and the deadline is set to $D=0.12$ s and $D=0.085$ s, respectively. It can be observed that the energy consumption keeps relatively high under the computing-prior and memory-prior policies, but the benefit of joint scaling policy is limited. This is because the available computing frequency is restricted by the frequency range, and the impact of computing frequency scaling on the inference time is not obvious according to Table \ref{tab:mytab1}. In conclusion, joint memory frequency and computing frequency scaling can provide more opportunities to reduce the energy consumption while meeting the deadline constraint.

\subsection{Cooperative Inference via Task Offloading}
\begin{figure}[t]
  \centering
  \includegraphics[width=0.4\textwidth]{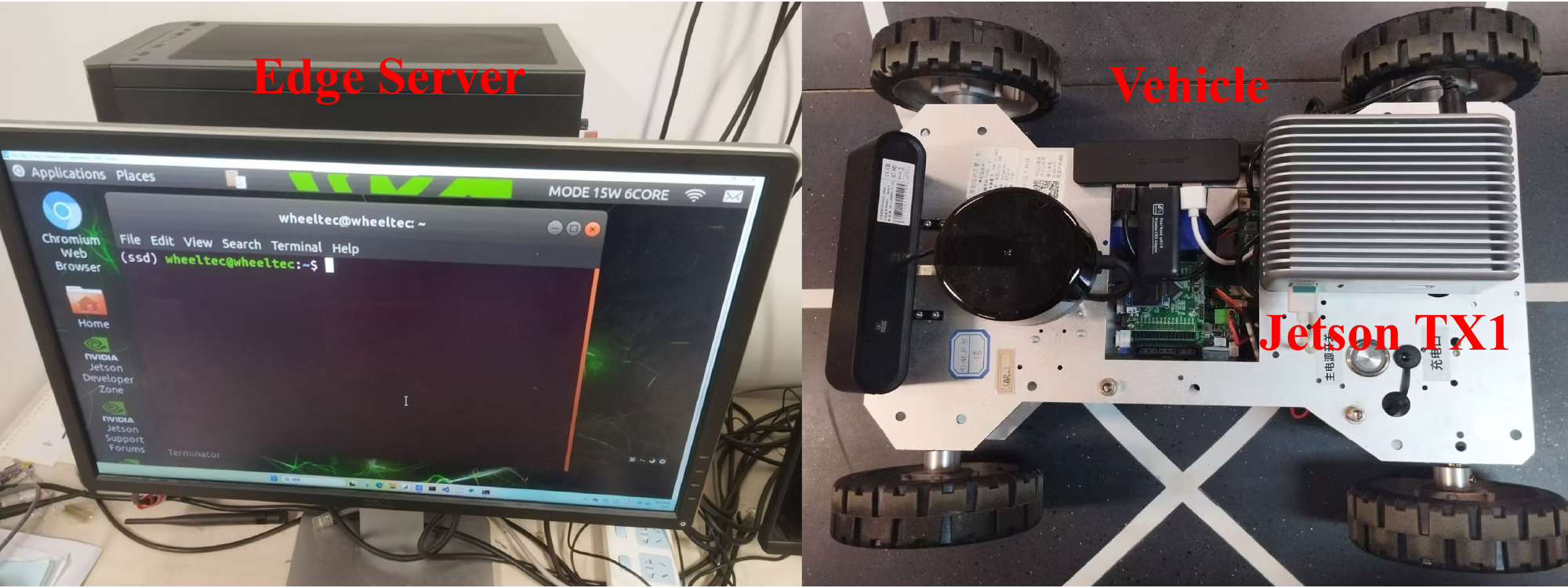}\\
  \caption{Test environment consisting of one vehicle equipped with Jetson TX1 and an edge server.}
  \label{fig:myFig9}
  \vspace{-0.25cm}
\end{figure}

As shown in Fig. \ref{fig:myFig9}, we consider a specific scenario consisting of one device (e.g., Jetson TX1) and an edge server (e.g., GeForce RTX 4080) to analyze the impact of memory frequency and computing frequency scaling. VGG19 and ResNet152 are deployed to complete the image classification task, and the input image size is $0.57$ MB. The transmit power is set to $0.2$ W. The binary offloading policy is adopted to achieve energy-efficient inference within the deadline.

\begin{figure}[t]
    \centering
    \subfigure[VGG19]{\includegraphics[width=0.4\textwidth]{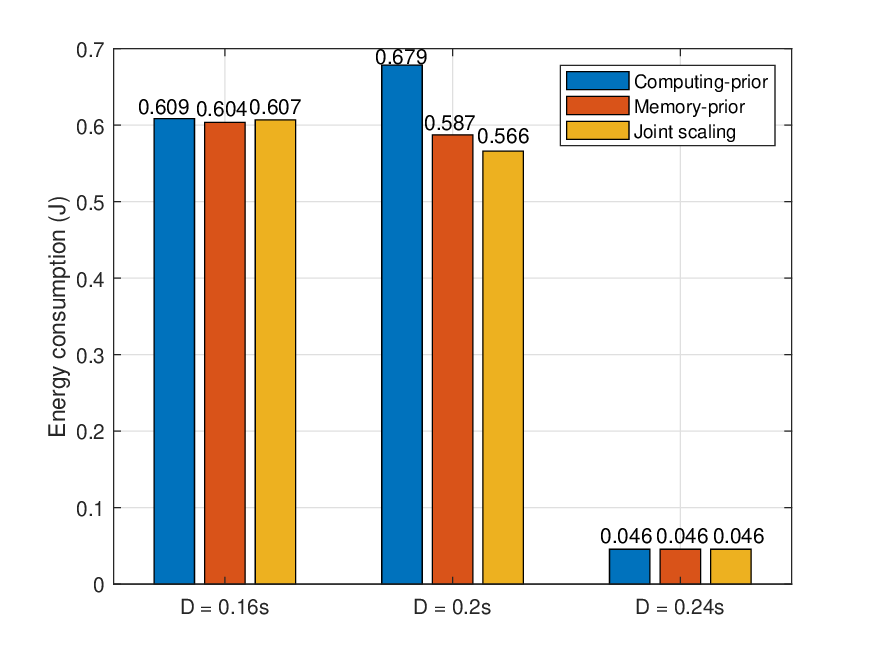}
    \label{fig:myfig10a}}
    \subfigure[DenseNet121]{\includegraphics[width=0.4\textwidth]{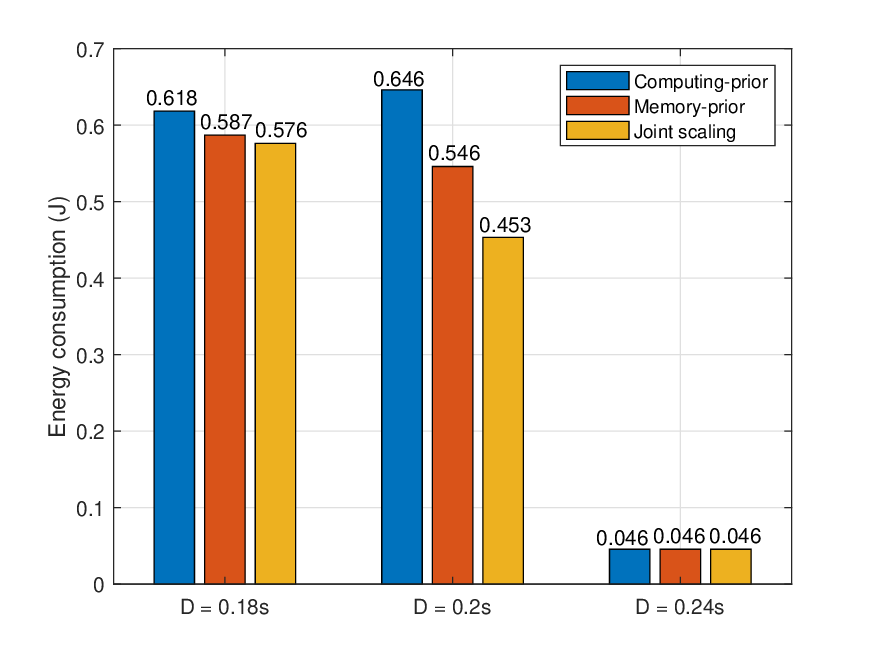}
    \label{fig:myfig10b}}
    \caption{Energy consumption of VGG19 and DenseNet121 under different deadlines.}
    \label{fig:myfig10}
    \vspace{-0.25cm}
\end{figure}

In Fig. \ref{fig:myfig10}, we analyze the total energy consumption of executing inference with VGG19 and DenseNet121 under different deadlines with the communication rate $R = 20$ Mbps. When the deadline is set to $D = 0.24$ s, which is larger than the sum of offloading delay and edge inference time, the device will transmit its data to the edge server to reduce the energy consumption. In this case, all policies lead to the same energy consumption, which is extremely small and results from the data transmission. As the deadline decreases, executing edge inference by offloading will not meet the constraint. Therefore, the device needs to execute local inference with memory frequency and computing frequency scaling to satisfy the deadline constraint, leading to a much larger energy consumption. Under the deadline $D = 0.2$ s, jointly scaling both memory frequency and computing frequency can reduce the energy consumption by an average of $23\%$ and $10\%$ compared with the computing-prior and memory-prior policy, respectively. However, the effects are no longer obvious when the deadline is further decreased to $D = 0.16$ s. This is because the deadline is close to the minimum inference time that the device can achieve, resulting in a finite frequency scaling range. In this case, the joint scaling policy brings limited gains and is not necessarily optimal. 

\section{Conclusion}\label{sec:conclusion}
In this paper, we have introduced the memory frequency and studied the impact of joint memory frequency and computing frequency scaling on DNN inference time and energy consumption. Considering the synergy between the two frequencies, we separately keep one frequency constant and change the other frequency to obtain the real-world inference time. With a model-based and data-driven method, a realistic model of the inference time and energy consumption is formulated. Additionally, an elementary analysis of the energy consumption is given to show the complex trade-off between memory frequency and computing frequency. Simulation results show that joint memory frequency and computing frequency scaling can reduce the energy consumption in two specific cases. 



\end{document}